# Coarse and Fine-Grained Hostility Detection in Hindi Posts using Fine Tuned Multilingual Embeddings

Arkadipta De*, Venkatesh E, Kaushal Kumar Maurya, and Maunendra Sankar Desarkar

**Indian Institute of Technology Hyderabad, Telangana, India**
{ai20mtech14002,ai20mtech14005,cs18resch11003}@iith.ac.in,
maunendra@cse.iith.ac.in

**Abstract.** Due to the wide adoption of social media platforms like Facebook, Twitter, etc., there is an emerging need of detecting online posts that can go against the community acceptance standards. The hostility detection task has been well explored for resource-rich languages like English, but is unexplored for resource-constrained languages like Hindi due to the unavailability of large suitable data. We view this hostility detection as a multi-label multi-class classification problem. We propose an effective neural network-based technique for hostility detection in Hindi posts. We leverage pre-trained multilingual Bidirectional Encoder Representations of Transformer (mBERT) to obtain the contextual representations of Hindi posts. We have performed extensive experiments including different pre-processing techniques, pre-trained models, neural architectures, hybrid strategies, etc. Our best performing neural classifier model includes *One-vs-the-Rest* approach where we obtained 92.60%, 81.14%, 69.59%, 75.29% and 73.01% F1 scores for hostile, fake, hate, offensive, and defamation labels respectively. The proposed model[1] outperformed the existing baseline models and emerged as the state-of-the-art model for detecting hostility in the Hindi posts.

**Keywords:** Neural Network · Hostility Detection · Transformer · Multilingual BERT

## 1 Introduction

The use of social media and various online platforms has increased drastically in recent times. A large number of users are engaged in social media platforms like - *Facebook*, *Twitter*, *Hike*, *Snapchat*, *Reddit*, *gab*, etc. The chat rooms, gaming platforms, and streaming sites are receiving a lot of attention. These fora are being increasingly used for discussions related to politics, governance, technology, sports, literature, entertainment etc. The law of freedom of speech [18] on social media has given the users the luxury to post, react, and comment freely, which generates a large volume of hostile contents too. In various circumstances these comments/posts are found to be biased towards a certain community, religion, or even a country. During the COVID-19 pandemic there has been around 200% increase[2] in traffic by hate and offensive speech promoters against the Asian community and a 900% increase in similar contents towards Chinese people. Around 70% increase in hate speech among teenagers and kids online, and a 40% increase in toxicity language by the gaming community has been reported. There have been many cases where hostile contents have led to incidents of violence (e.g., mob-lynching), communal riots, racism, and even deaths across the world. Hence there is a need to detect and prevent such activities in online fora. This is the major motivation for the task of Hostile post detection. More specifically, we aim to detect hostile content in Hindi posts.

---

* Corresponding Author
[1] https://github.com/Arko98/Hostility-Detection-in-Hindi-Constraint-2021

There are many recent work for hostility detection such as hate speech detection on Twitter, for posts written in English [6,1,20]. Although Hindi is the third most spoken language in the world, it is considered as a resource-poor language. Hindi sentences have diverse typological representations as compared to English. Due to these facts, multiple challenging NLP problems including hostility detection are still unexplored for Hindi-language text. We tackle the hostility detection problem in Hindi posts as a two-step process: First, we employ *Coarse-grained Classification* to identify *Hostile* or *Non-Hostile* contents. Secondly, we further classify the hostile posts into four fine-grained categories, namely, *Fake*, *Hate*, *Defamation*, and *Offensive* through *Fine-grained Classification*. In summary, the problem can be viewed as a *multi-label multi-class classification* problem. The definitions [14,19] of different class labels are included below:

1. **Fake News**: A claim or information that is verified to be not true. Posts belonging to clickbait and satire/parody categories can be also categorized as fake news.
2. **Hate Speech**: A post targeting a specific group of people based on their ethnicity, religious beliefs, geographical belonging, race, etc., with malicious intentions of spreading hate or encouraging violence.
3. **Offensive**: A post containing profanity, impolite, rude, or vulgar language to insult a targeted individual or group.
4. **Defamation**: A misinformation regarding an individual or group, which is destroying their reputation publicly.
5. **Non-Hostile**: A post with no hostility.

We propose a multilingual BERT based neural model that outperformed the existing baselines and emerged as the state-of-the-art model for this problem. We perform extensive experiments including multiple pre-processing techniques, pre-trained models, architecture exploration, data sampling, dimension reduction, hyper-parameter tuning, etc. The detailed experimental analysis and discussions provide insights into effective components in the proposed methodology for the task at hand. The rest of the paper is organized as follows: Related literature for hostility detection is presented in Section 2. The methodology is discussed in Section 3; Section 4 presents the experimental setup. Section 5 presents the experimental evaluations, and we conclude our discussion in Section 6.

---

[2] https://l1ght.com/Toxicity_during_coronavirus_Report-L1ght.pdf

## 2 Related Works

Here, we briefly review existing works from the literature on hostility detection.

• **Hostility Detection in the English Language:** English being the most widely adopted language on social media platforms, several notable works exist for hostility detection in the English language. A comprehensive review of detecting fake news on social media, including fake news characterizations on psychology and social theories is presented in [17]. Ruchansky et. al. [16] consider *text, response* and *source* of a news in a deep learning framework for fake news detection. In [12], the authors propose methods to combine information from different available sources to tackle the problem of Multi-source Multi-class Fake-news Detection. A lexicon-based approach is proposed by [7] to hate speech detection in web discourses viz. web forums, blogs, etc. Djuric et. al. [6] propose distributed low-dimensional representation based hate speech detection for online user comments. A deep learning architecture to learn semantic word embeddings for hate speech detection is presented in [1].

• **Hostility Detection in Non-English Languages:** In [8], the authors address the problem of offensive language detection in the Arabic language using Convolution Neural Network (CNN) and attention-based Bidirectional Gated Recurrent Unit (Bi-GRU). A novel dataset of $50k$ annotated fake news in Bengali language is released in [9]. A fastText-based model has been used by [11] for the classification of offensive tweets in the Hindi language written in Devanagari script. The authors also release an annotated dataset for the detection of Hindi language abusive text detection. Bohra et. al. [3] analyzed the problem of detecting hate speech in Hindi-English code-mixed social media text. They proposed several classifiers for detecting hate speech based on a sentence level, word level, and lexicon-based features.

Unlike previous works, we propose an approach based on transformer's encoder based pre-trained multilingual models with multiple neural architectures to detect hostility in Hindi posts. The work has been conducted as a part of Shared task at CONSTRAINT 2021 Workshop [15] as IITH-BRAINSTORM team.

## 3 Methodology

In this section, we present our proposed models for coarse-grained and fine-grained tasks of hostility detection in Hindi posts. The backbone of our proposed model is Transformer's encoder based pre-trained architecture BERT [5]. More specifically, we leverage the multi-lingual version of BERT (mBERT) [5] and XLM-Roberta [4]. XLM-Roberta is a variant of BERT with a different objective, and is trained in an unsupervised manner on a multi-lingual corpus. These models have achieved state-of-the-art results in NLU and NLG tasks across multiple languages for popular benchmarks such as XGLUE [13], XTREME [10].

### 3.1 Coarse-Grained Classification

These sections include details of the models which were used for a coarse-grained classification task.

• **Fine-Tuned mBERT (FmBERT) and XLM-R (FXLMR) Models:** For the coarse-grained task we fine-tune the mBERT (*bert-base-multilingual-cased*) and XLM-Roberta (*xlm-roberta-base*) models for the binary classification problem (i.e., hostile or non-hostile). An architectural diagram of the model is shown in Figure 1a. In fine-tuning phase, for each post we use last layer *[CLS]* token representation (a 768-dimensional vector) from mBERT/XLM-Roberta.

• **Coarse Grained Hybrid Model (CoGHM):** To further improve the performance of the Coarse-grained classification task, we propose a model that combines representations from mBERT and XLM-Roberta. We obtain the last layer hidden representation from the two models and concatenate them. The concatenated representation is fed through a three-layer MLP (Multi-layered Perceptron) model. Subsequently, softmax operation has been applied to the MLP output to obtain the class labels (see Figure 1b).

• **Recurrent Neural Models:** We also explore Long Short Term Memory (LSTM) and Gated Recurrent Unit (GRU) based neural network architectures to observe their performances on the task. These models are known to capture long term dependencies. We took each sub-word representation (extracted features of given Hindi post) of mBERT and pass them to the Bidirectional versions of LSTM or GRU (i.e, BiLSTM or BiGRU) layers. Hidden representations from these models are passed through an MLP (with 3 layers) and softmax layer to obtain the final class labels (see figure 1c).

• **Traditional Machine Learning Models:** To observe the behaviour of traditional machine learning models we performed experiments with widely popular algorithms such as Support Vector Machine (SVM), Random Forest (RF), and Gradient Boosted Decision Tree (GBDT). For a given post, we extracted each subword representation from mBERT model. The dimension of each post is now $m \times 768$, where $m$ is the number of subwords in the post. We also applied Principal Component Analysis (PCA) to each sub-word representation to reduce its dimension from 768 to 20. After concatenating the reduced representations of the sub-words of the post, the concatenated representation is fed through the above classification algorithms. The model diagram is shown in Figure 1d.

### 3.2 Fine-Grained Classification

The fine-grained classification deals with further categorizing the hostile posts into specific sub-categories such as *Fake, Hate, Offensive*, and *Defamation*.

• **Direct Multi-label Multi-classification (DMLMC) Model:** In this setting, we adopted standard multi-label multi-class classification architecture with pre-trained contextual sentence embedding from mBERT/XLM-R. First, we extract the sentence representation of each Hindi post from mBERT/XLM-R (i.e., $[CLS]$ token representation) and pass it through a 3 layered MLP model to obtain the representation $h_1$. Finally, $h_1$ is passed though a Sigmoid layer with 4 independent neurons. The output of the Sigmoid layer is a $1 \times 4$ dimensional vector $p$ where each cell corresponds to an *independent probability* of the post belonging to the four hostile classes. While training this module, we consider only the hostile instances (i.e instances annotated as *Fake*, *Hate*, *Offensive*, or *Defamation*). The architectural diagram of this model is shown in Figure 1e.

• **One vs Rest (OvR) Model:** In this setting, we reformulate the multi-class classification problem as four separate binary classification problems. For each class, there is a separate classifier that is trained independently. Predictions of the individual classifiers are merged to obtain the final multi-label prediction. For each model, we take the 768-dimensional pooled representation from mBERT model and feed them to 3-layered MLP. The output representation from the MLP layer is passed through to a softmax layer to get the final classification label. The architecture diagram is given in Figure 1f.

The primary difference between DMLMC and OvR model architectures lies in the training data and procedure. OvR builds four different models with four binary classification datasets (Hate vs Non-Hate, Fake vs Non-fake, etc.), and each model gives a *Yes* or *No* response. Binarization of a particular class has been done by assigning *Yes* to instances annotated as belonging to that particular class, and *No* for all other instances that was marked as belonging to other hostile classes in the original dataset. This process has been done for all the four hostile classes. On the other hand, the DMLMC model is trained with a single dataset where posts are labeled as *Fake*, *Hate*, *Offensive*, or *Defamation*.

## 4 Experimental Setup

• **Dataset:** We use the Hindi hostile dataset proposed in [2] containing 8200 hostile and non-hostile posts from Facebook, Twitter, Whatsapp, etc. Each post is annotated by human annotators as *Non-hostile* or *Hostile*. Further, hostile posts are annotated with fine-grained labels such as *Fake*, *Hate*, *Defamation*, and *Offensive*. The *Fake-news* related data was collected from India's topmost fact-checking websites like BoomLive[3], Dainik Bhaskar[4], etc. Other posts of the dataset were collected from popular social media platforms. A brief statistics and sample data instances are shown in Tables 1 and 2 respectively. It can be noticed that a particular data instance can have multiple hostile labels.

• **Preprocessing:** We perform several pre-processing steps on the dataset. Pre-processing steps include removal of non-alphanumeric characters (i.e., @, _, $ etc.), emoticons (i.e., *:-)*, *:-(*, etc.), newline and new paragraph characters. Additionally, we also experimented with removing stop-words, removing NERs and performing stemming.

---

[3] https://hindi.boomlive.in/fake-news
[4] https://www.bhaskar.com/no-fake-news/

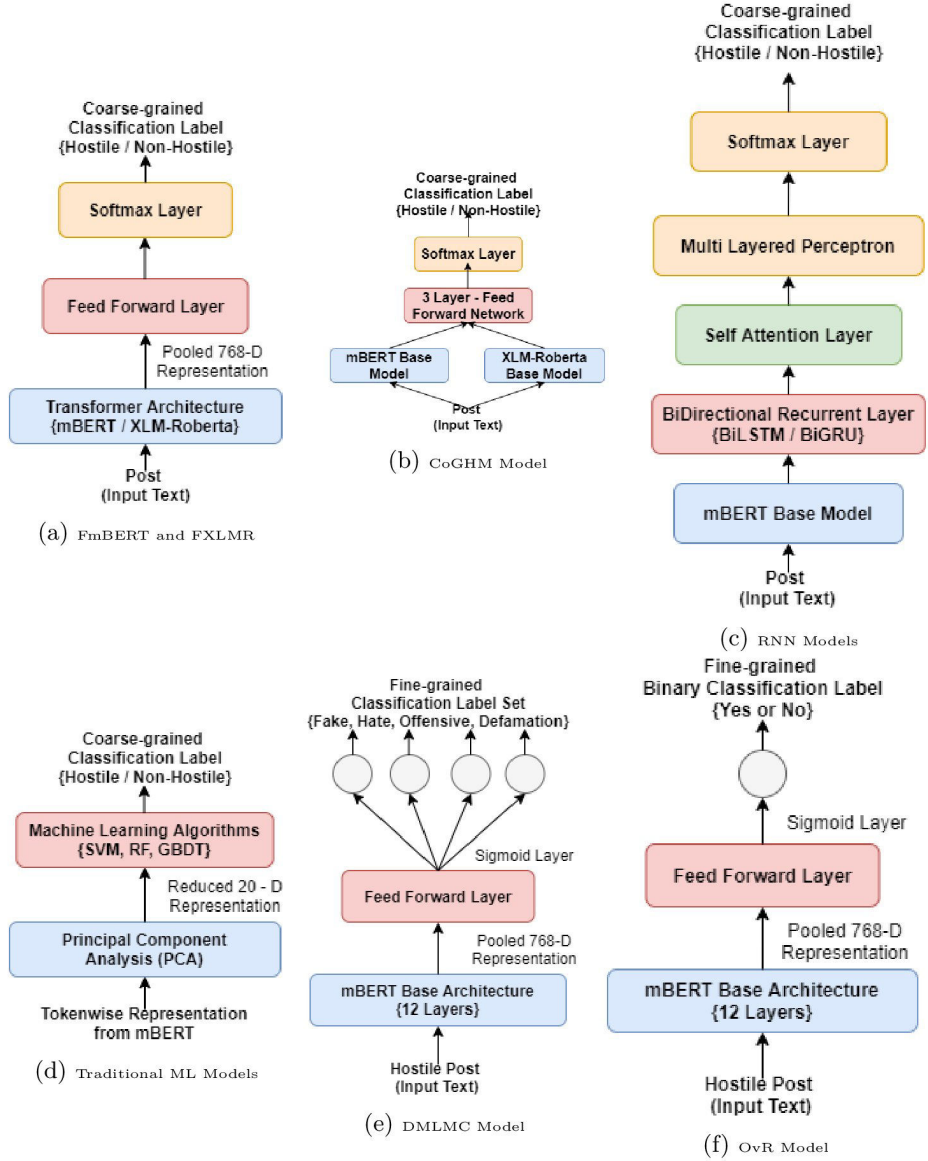

Fig. 1: Architecture Diagrams for Coarse-grained and Fine-grained Evaluation

- **Baseline Models and Evaluation Metrics:** We have included baseline models from [2] which is the data source paper. They extracted the last layer $[CLS]$ token representation from mBERT and fed that as input to traditional machine learning algorithms like SVM, Logistic Regression, Random Forest, and Multi-Layer Perceptron. Similar to the baseline paper, Accuracy, and Weighted

Hostility Detection in Hindi Posts using Multilingual Embeddings        7| Category | Hostile | | | | Non Hostile |
|---|---|---|---|---|---|
| | Fake | Hate | Defame | Offense | |
| Train | 1144 | 792 | 742 | 564 | 3050 |
| Dev | 160 | 103 | 110 | 77 | 435 |
| Test | 334 | 237 | 219 | 169 | 873 |
| Total | 1638 | 1132 | 1071 | 810 | 4358 |

Table 1: Dataset Statistics

| Sl. | Post | Labels |
|---|---|---|
| 1 | मेरे देश के हिन्दु बहुत निराले है। कुछ तो पक्के राम भक्त है और कुछ बाबर के साले है जय श्री राम | Hate, Offensive |
| 2 | JEE Exam center से निकले #Students को सुन बाकी छात्रों के साथ Parents के चेहरे पर मुस्कान आ जाएगी https://t.co/TQ7nfIv0I0 https://t.co/gGCDYYEz6E | Non-Hostile |
| 3 | कांग्रेस मूल की कंगना रनौत बिहार चुनाव में भाजपा का प्रचार करेंगी! #NATIONALNEWS | Fake |
| 4 | @SalmanNizami_ राहुल गांधी – Maa मैं अगले 4 साल क्या करूंगा सोनिया गांधी – बेटा TV रिचार्ज कर दिया है बैठकर छोटा भीम देख. | Defamation |

Table 2: Example of Dataset

Average F1-Score are used as primary evaluation metrics for coarse-grained and fine-grained evaluation respectively.

• **Implementation Details:** We set the maximum input sequence length to 128, *Warmup* proportion to 0.15, batch size to 28, and number of epoch to 10. For mBERT and XLM-Roberta models, we use an initial learning rate of 2E-5 and 5E-5 respectively. We use GeLU as a hidden activation function and use 10% Dropout. Other parameters of mBERT[5] and XLM-Roberta[6] are not modified. We adopted grid search to find the best performing set of hyper-parameters. SVM uses Gaussian kernel (RBF kernel) and the number of estimators for Random Forest is set to 80. For LSTM and GRU, 2 recurrent layers are used.

## 5 Results and Discussion

### 5.1 Coarse-grained Evaluation

Table 3 compares the results of directly fine-tuned models *FmBERT* and *FXLMR*, the hybrid model *CoGHM*, and traditional machine learning-based models (SVM, RF, GBDT, and XGBoost (with and without PCA).

We obtain **91.63%** and **89.76%** accuracy scores on direct fine-tuning mBERT and XLM-Roberta models respectively on the binary classification objective. The hybrid model (CoGHM) has an accuracy score **92.60%** and emerges as our best performing model for coarse-grained evaluation. BiLSTM and BiGRU have similar scores compared to CoGHM, which indicates the effectiveness of the two architectures. As shown in Table 5, accuracies of *FmBERT* and *FXLMR* models drop if Named Entities are removed or stemming is performed. This observation indicates that every piece of information is crucial in the online posts due to its non-traditional sentence structure. The confusion matrix of the CoGHM model on validation data is given in Figure 2a. For traditional machine learning models, XGBoost performed better than others, but there is no significant difference observed across these models. A similar situation is observed with and without PCA with 20 dimensions. This shows that the embeddings learned by the transformer models capture different non-overlapping aspects, and are representative enough for discriminating the hostile posts from non-hostile ones.

---

[5] https://github.com/google-research/bert/blob/master/multilingual.md
[6] https://github.com/pytorch/fairseq/tree/master/examples/xlmr

| Algorithm | PCA | Accuracy (%) |
|---|---|---|
| FmBERT | - | 91.63 |
| FXLMR | - | 89.76 |
| CoGHM | - | **92.60** |
| BiLSTM | - | 92.11 |
| BIGRU | - | 92.36 |
| SVM | Yes | 91.86 |
| | No | 91.49 |
| RF | Yes | 91.61 |
| | No | 91.46 |
| GBDT | Yes | 91.63 |
| | No | 91.46 |
| XGBoost | Yes | 91.98 |
| | No | 91.62 |

Table 3: Coarse-grained evaluation results with multilingual pre-trained models

| Hostile Label | DMLMC mBERT | DMLMC XLMR | OvR |
|---|---|---|---|
| Fake | 51.06 | 53.72 | 81.14 |
| Hate | 56.91 | 60.11 | 69.59 |
| Defame | 59.57 | 57.97 | 73.01 |
| Offense | 64.89 | 67.77 | 75.29 |
| Average | 30.00 | 32.88 | 69.57 |

Table 4: Weighted F1 score for Fine-grained Evaluation

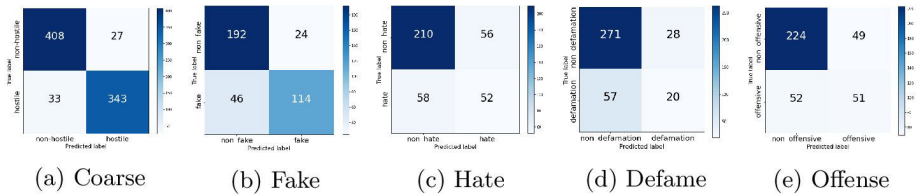

(a) Coarse   (b) Fake   (c) Hate   (d) Defame   (e) Offense

Fig. 2: Confusion Matrix of Coarse-grained CoGHM and Fine-grained Evaluation for One vs Rest Approach.

### 5.2 Fine-grained Evaluation

| Model | Pre-pro | Accuracy (%) |
|---|---|---|
| FmBERT | NE-Rem | 91.49 |
| | Stemmed | 91.63 |
| | NE-Rem Stemmed | 90.64 |
| FXLMR | NE-Rem | 89.04 |
| | Stemmed | 89.76 |
| | NE-Rem Stemmed | 88.57 |

Table 5: Result of Coarse-grained Models with pre-processing strategies (NE-Rem: Results after removing Named Entities from text, Stemmed: Results with stemmed tokens in text)

| Model | Coarse Grained | Fine Grained | | | |
|---|---|---|---|---|---|
| | | Fake | Hate | Offense | Defame |
| LR | 83.98 | 44.27 | **68.15** | 38.76 | 36.27 |
| SVM | **84.11** | **47.49** | 66.44 | **41.98** | **43.57** |
| RF | 79.79 | 6.83 | 53.43 | 7.01 | 2.56 |
| MLP | 83.45 | 34.82 | 66.03 | 40.69 | 29.41 |
| Ours | **92.60** | **81.14** | **69.59** | **75.29** | **73.01** |

Table 6: Comparison of baseline with best proposed model for Coarse-grained (Accuracy) and Fine-grained (f1 score) evaluation on Validation Data

In fine-grained evaluation, the average F1 score is computed across the hostile classes. The results for DMLMC and OvR models are shown in Table 4. OvR model performed significantly better as compared to the DMLMC model across all the labels. In the OvR method, features that are important and contribute more towards a specific class are not suppressed by features that are important for other classes. It may be the case that, some features that positively contribute towards the classification of a particular class negatively contribute towards the classification of other class. Even in that case, the subword gets its class-specific proper importance in the OvR method. Figure 2 shows the confusion matrix for the OvR model.

### 5.3 Comparison with Baseline

We compare our proposed model's performance with baseline [2] models in Table 6. We can observe that our proposed model performs better for both coarse-grained and fine-grained evaluation. The performance margin for Coarse-grained evaluation is 8.49% and for fine-grained evaluation, the maximum margin was 33.65% on *Fake* posts. For the hate category, we have received comparatively poor performance gain (the margin is 1.44%). A possible reasoning could be that the hate posts are semantically similar to the other labels of hostile data and the model got confused during training and prediction for this class.

### 5.4 Additional Discussions and Analysis

In this section, we present a brief study of our best model's predictions on validation data and discuss important observations.

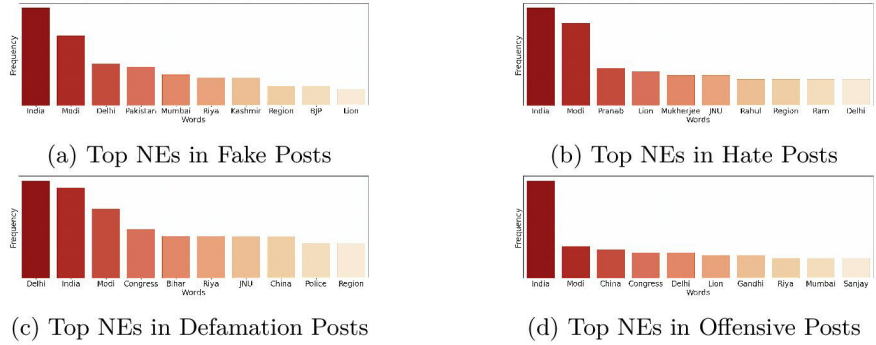

(a) Top NEs in Fake Posts

(b) Top NEs in Hate Posts

(c) Top NEs in Defamation Posts

(d) Top NEs in Offensive Posts

Fig. 3: Top Frequent Named Entities in different categories of Predicted Posts

1. From the frequency plots in Figure 3 we observed that the words "**India**" and "**Modi**" are the top frequent words in posts classified as *Fake*, *Hate*, *Offensive*, and *Defamation*. This gives us a clear indication that a lot of *Hostile* sentences are regarding politics as political NEs like "**Modi**", "**Rahul**" are predominantly present in the Hostile posts.
2. Specific words like "**Congress**" are associated with the classes *Offensive* and *Defamation*, whereas the word "**Pakistan**", "**Delhi**" are associated with *Fake* posts and the word "**JNU**" is associated with *Hate* speech.
3. We also observe that current events (such as the Corona Virus outbreak, death of a Bollywood actor, JNU attack, etc.) have a very important role in deciding posts to be detected as Hostile. Example - the association of word "**China**" and "**Riya**" with *Offensive* and *Defamation* posts.
4. By examining further we also observe that presence of the words like "**RSS**", "**Ram**", "**Kashmir**", "**Region**" etc. increases the probability of a post being classified as *Offensive* and *Hate*.
5. It is also observed that the probability of a sentence being classified as *Offensive* increases very sharply if the post contains vulgar words such as "साले" "कुत्ते" etc.

## 6    Conclusion and Future Works

In this paper, we tackle the important and relevant problem of "detection of hostile Hindi posts". We have performed extensive experiments with multiple models, and architectures with different representations of the input texts. Our one-vs-rest neural classifier on top of mBERT neural representations of posts emerged as the best performing model. In future, we would like to extend the work to low resource languages other than Hindi such as Vietnamese, Indonesian, Telugu, Tamil, Swahili, etc. Investigating the effect of considering different linguistic features to detect hostility in different posts will be an interesting research direction.